\documentclass{article} % For LaTeX2e
\usepackage{iclr2018_conference,times}
\usepackage{hyperref}
\usepackage{url}
\usepackage{amsmath,amssymb,bm,amsfonts}
\usepackage{graphicx}
\newcommand*{\QEDA}{\hfill\ensuremath{\blacksquare}}
\newtheorem{theorem}{Theorem}

\title{Topology Adaptive Graph Convolutional \\ Networks  }

\author{Jian Du, Shanghang Zhang, Guanhang Wu,  Jos{\'e} M. F. Moura \& Soummya Kar\\
	Carnegie Mellon University, Pittsburgh, PA\\
	\texttt{\{jiand,shanghaz,guanhanw,moura,soummyak\}@andrew.cmu.edu} \\
}

\iclrfinalcopy % Uncomment for camera-ready version, but NOT for submission.

\begin{document}

\maketitle

\begin{abstract}
	%Convolution  acts as a local feature extractor in  convolutional neural networks (CNNs).
	%However, the convolution operation is not applicable when the input data  is supported on an irregular graph such as with social networks, citation networks, or knowledge graphs.
	Spectral graph convolutional neural networks (CNNs) require approximation to the convolution to alleviate the computational complexity, resulting in performance loss. This paper proposes the topology adaptive graph convolutional network (TAGCN), a novel graph convolutional network defined in the vertex domain.
	We provide a systematic way to design a set of fixed-size learnable filters to perform convolutions on graphs.
	The topologies of these filters are adaptive to the topology of the graph when they
	scan the graph to perform convolution.
	The  TAGCN not only inherits the properties  of convolutions in CNN for grid-structured data, but it is also consistent
	with  convolution  as defined in graph  signal processing.
	{Since  no approximation to the convolution is needed, TAGCN exhibits  better performance than existing spectral CNNs on a number of data sets and is also   computationally simpler than other  recent methods.}
\end{abstract}

\section{Introduction}
Convolutional neural network (CNN) architectures exhibit state-of-the-art performance on a variety of learning tasks dealing with 1D, 2D, and 3D grid-structured data such as acoustic signals, images, and videos, in which
convolution   serves as a feature extractor~\cite{lecun2015deep}. However, the (usual) convolution operation is not applicable when  applying CNN  to data that is  supported on an arbitrary graph rather than on a regular grid structure, since the number and topology of neighbors of each vertex on the graph varies, and  it is difficult to design a fixed-size filter scanning over the graph-structured data for feature extraction.

Recently, there has been an increasing interest in graph CNNs~\cite{bruna2013spectral, defferrard2016convolutional, thomas2017semi,monti2017geometric, levie2017cayleynets}, attempting to generalize deep learning methods to graph-structured data, specifically focusing on the design of graph convolution.
In this paper, we propose the topology adaptive graph convolutional network (TAGCN), a unified convolutional neural network to learn nonlinear representations for the graph-structured data. It slides a set of fixed-size learnable filters on the graph simultaneously, and the output is the weighted sum of these filters' outputs, which extract both vertex features and strength of correlation between vertices. Each filter is adaptive to the topology of the local region on the graph where it is applied. TAGCN unifies filtering in both the spectrum and vertex domains; and applies to both directed and undirected graphs.
%Various previous graph CNN can be considered as particular instances of our framework.

In general, the existing graph CNNs can be grouped into two types: spectral domain techniques and vertex domain techniques. In \citet{bruna2013spectral}, CNNs have been generalized to   graph-structured data, where convolution is achieved by a pointwise product in the spectrum domain according to the convolution theorem.
Later, \citet{defferrard2016convolutional} and \citet{levie2017cayleynets} proposed   spectrum filtering based methods that utilize  Chebyshev polynomials and Cayley polynomials, respectively. 
\citet{thomas2017semi} simplified this spectrum method and obtained a filter in the vertex domain, which achieves   state-of-the-art performance.
Other researchers~\citep{diffusion, monti2017geometric} worked on designing feature propagation models in the vertex domain for graph CNNs.
\citet{yang2016revisiting, dai2016discriminative, grover2016node2vec, du2016convergence}  study transforming graph-structured  data to embedding vectors for  learning problems.
More recently, \citep{Bengio18} proposed graph attention networks leveraging masked self-attentional layers to address the approximation of exiting graph convolutions networks.
Nevertheless, it still remains open how to extend CNNs from grid-structured data to arbitrary graph-structured data with local feature extraction capability.

%Our proposed TAGCN  is graph-based convolution and draws on techniques from graph signal processing. 
We define rigorously the graph convolution operation on the vertex domain as multiplication by polynomials of the graph adjacency matrix, which is consistent with the notion of convolution in graph signal processing.
In graph signal processing \cite{sandryhaila2013discrete}, polynomials of the adjacency matrix are graph filters, extending to graph based data from the usual concept of filters in traditional time or image based signal processing. Thus, comparing ours with existing work on graph CNNs, our paper provides a solid theoretical foundation for our proposed convolution step instead of an ad-hoc approach to convolution in CNNs for graph structured data. 

{Further, our method avoids computing the spectrum of the graph Laplacian as in \citet{bruna2013spectral}, or approximating the spectrum using high degree Chebyshev polynomials of the graph Laplacian matrix (in \citet{defferrard2016convolutional}, it is suggested that one needs a $25^{\textrm{th}}$ degree Chebyshev polynomial to provide a good approximation to the graph Laplacian spectrum) or using high degree Cayley polynomials of the graph Laplacian matrix (in \citet{levie2017cayleynets}, $12^{\textrm{th}}$ degree Cayley polynomials are needed). We also clarify that the GCN method in \citet{thomas2017semi} is a first order approximation of the Chebyshev polynomials approximation in \citet{defferrard2016convolutional}, which is very different from our method. Our method has a much lower computational complexity than the spectrum based methods, since our method only uses polynomials of the adjacency matrix with maximum degree $2$ as shown in our experiments. Finally, the method that we propose exhibits better performance than existing methods as no approximation is required.}
Our contributions are summarized as follows:
\begin{itemize}
	\item
	The proposed TAGCN explores a general $K$-localized filter for graph convolution in the vertex domain to extract  local features on a set of size-$1$ up to size-$K$ receptive fields.
	The topologies of these filters are adaptive to the topology of the graph  as they scan  the graph to perform convolution.
	It replaces the fixed   square filters in traditional CNNs for the  grid-structured input data volumes in  traditional CNNs.
	Thus, the convolution  that we define in the convolution step for the vertex domain is consistent with convolution in traditional  CNNs.
	\item
	{We analyze the mechanisms of the graph convolutional layers and prove that if only a size-k filter is used, as the convolutional layers go deeper under certain condition, the output of the last convolutional layer is the projection of the output of the first convolutional layer  along the eigenvector corresponding to the  eigenvalue of the graph adjacency matrix with the largest amplitude.
		This linear approximation leads to information loss and classification accuracy degradation.
		In contrast, using a set of size-1 up to size-K filters (as in our TAGCN) can avoid the linear approximation and increases the representation capability. Therefore, it leads to improved classification accuracy.}
	\item
	{TAGCN  is  consistent with  the convolution  in graph signal processing.
		It  applies to both directed and undirected graphs.
		Moreover, it has a much lower computational complexity compared with recent methods  since it only needs polynomials of the adjacency matrix with maximum degree $2$ compared with the  $25^{\textrm{th}}$ and $12^{\textrm{th}}$ degree Laplacian matrix polynomials  in \citet{defferrard2016convolutional} and \citet{levie2017cayleynets}.  }
	
	\item
	{As no approximation to the convolution is needed in TAGCN, it achieves better performance compared with existing methods.}
	We contrast TAGCN with  recently proposed  graph CNN  including both   spectrum filtering methods \citep{bruna2013spectral,defferrard2016convolutional} and vertex domain propagation methods \citep{thomas2017semi,monti2017geometric,diffusion},
	evaluating their performances on three commonly used graph-structured data sets.
	Our experimental tests show that   TAGCN consistently achieves superior performance on  all these data sets.
\end{itemize}

\section{Convolution on Graph}
We use boldface uppercase and lowercase letters to represent matrices and vectors, respectively.
The information and their relationship on a graph $\mathcal{G}$ can be represented by
$\mathcal{G} = (\mathcal V, \mathcal {E}, \bar{\textbf A})$, where $\mathcal V$ is the set of vertices, $\mathcal E$ is the set of edges, and $\bar{\textbf A}$ is the weighted  adjacency matrix of the graph; the graph can be weighted or unweighted, directed or undirected.
We assume there is no isolated vertex in $\mathcal G$.
If $\mathcal{G}$ is a \emph{directed weighted} graph, the weight $\bar{\textbf A}_{n,m}$ is on the directed edge from vertex $m$ to $n$.
The entry  $\bar{\textbf A}_{n,m}$ reveals the dependency between node $n$ and $m$ and  can take arbitrary real or  complex values.
The  graph convolution  is general and can be adapted to graph CNNs  for particular tasks.
In this paper, we focus on  the vertex semisupervised learning problem,
where we have access to very limited labeled vertices, and the task is to classify the remaining unlabeled vertices by feeding the output of the last convolutional layer to a fully connected layer.

\subsection{Graph Convolutional Layer for TAGCN}
Without loss of generality,  we  demonstrate   graph convolution on the $\ell$-th hidden layer. The results apply to any other hidden layers.
Suppose on the $\ell$-th hidden layer, the input feature map for each vertex of the graph has $C_{\ell}$ features.
We collect the $\ell$-th hidden layer
input data on all vertices
for  the $c$-th feature by the vector  $\textbf x^{(\ell)}_{c}\in \mathbb R^{N_{\ell}}$, where
$c = 1, 2,\ldots C_{\ell}$ and $N_{\ell}$ is the number of vertices\footnote{Graph coarsening could be used and the number of vertices may vary for different layers.}.
The components of $\textbf x^{(\ell)}_{c}$ are indexed by vertices of the data graph representation
$\mathcal G=(\mathcal V, \mathcal {E}, \bar{\textbf A})$\footnote{We use superscript $(\ell)$ to denote data on the $\ell$th layer and superscript $ {\ell}$ to denote the $\ell$-th power of a matrix.}.
Let $\textbf G^{(\ell)}_{c,f}\in \mathbb R^{N_{\ell}\times N_{\ell}}$ denote the $f$-th graph filter.
The graph convolution is the matrix-vector product, i.e., $\textbf G^{(\ell)}_{c,f}\textbf x^{(\ell)}_{c}$.
Then  the $f$-th output feature map followed by a ReLU function is given by
\begin{equation}\label{out_f}
\textbf y_f^{(\ell)} = \sum_{c=1}^{C_{\ell}}\textbf G^{(\ell)}_{c,f}\textbf x^{(\ell)}_{c} +  b_f\textbf 1_{N_{\ell}},
\end{equation}
where $ b_f^{(\ell)}$ is a learnable bias, and
$\textbf 1_{N_{\ell}}$ is the $N_{\ell}$ dimension  vector of all ones.
We   design  $\textbf G^{(\ell)}_{c,f}$ such that $\textbf G^{(\ell)}_{c,f}\textbf x^{(\ell)}_{c}$ is a meaningful convolution on a graph with arbitrary topology.

In the recent theory on graph signal processing~\citep{sandryhaila2013discrete}, the \emph{graph shift} is defined as a local operation that replaces a graph signal at a graph vertex by a linear weighted combination of the values of the graph signal at the neighboring vertices:
$$\tilde{\textbf x}^{(\ell)}_{c} = \bar{\textbf A} \textbf x^{(\ell)}_{c}.$$
The graph shift $\bar{\textbf A}$ extends the time shift in traditional signal processing to graph-structured data.
Following \citet{sandryhaila2013discrete},
a graph filter $\textbf G^{(\ell)}_{c,f}$ is shift-invariant, i.e.,
the shift $\bar{\textbf A}$ and the filter
$\textbf G^{(\ell)}_{c,f}$ commute,
$\bar{\textbf A}(\textbf G^{(\ell)}_{c,f}\textbf x_c^{(\ell)}) =
\textbf G^{(\ell)}_{c,f} (\bar{\textbf A} \textbf x_c^{(\ell)})$,  if under appropriate assumption $\textbf G^{(\ell)}_{c,f}$ is a polynomial in $\textbf A$,
\begin{equation}\label{con}
\textbf G^{(\ell)}_{c,f} =  \sum_{k = 0}^{K} g^{(\ell)}_{c,f,k}\textbf A^{k}.
\end{equation}
In (\ref{con}),
the $g^{(\ell)}_{c,f,k}$ are the graph filter polynomial coefficients; the quantity
$\textbf A = \textbf D^{-\frac{1}{2}}\bar{\textbf A}\textbf D^{-\frac{1}{2}}$ is the normalized adjacency matrix
of the graph, and $\textbf D = \text{diag}[\textbf d]$ with the $i$th component being $\textbf d(i) = \sum_j \textbf A_{i,j}$.\footnote{There is freedom to  normalize $\textbf A$ in different ways; here it is assumed that $\bar{\textbf A}_{m,n} $ is nonnegative and the above normalization is well defined.}
We adopt the normalized  adjacency matrix  to guarantee that
all the eigenvalues of $\textbf A$  are inside  the unit circle, and therefore $\textbf G^{(\ell)}_{c,f}$ is computationally stable.
The next subsection shows we will adopt   $  1\times C_{\ell},   2\times C_{\ell},\ldots,$ and $  K\times C_{\ell}$ filters sliding on the graph-structured data.
This fact coincides with  GoogLeNet \citep{googlenet}, in  which a set of  filters with different sizes are used in each convolutional layer.
Further, It is shown in the appendix that the convolution operator defined in (\ref{con}) is consistent with  that in
classical  signal processing when the graph is in the 1D cyclic  form, as shown in Fig.~\ref{f1}.
\begin{figure}[ht]
	\begin{center}
		\includegraphics[width=0.4\columnwidth, height = 1.6cm]
		{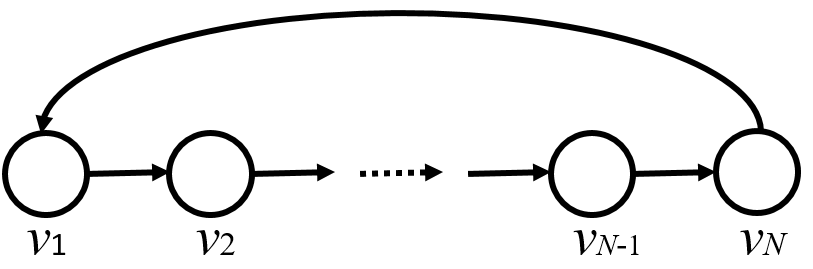}
	\end{center}
	\caption{Graph topology of a 1-D cyclic graph.}
	\label{f1}
\end{figure}

%By collecting all the   $C_{\ell}$ channels for layer $\ell$, we obtain $\textbf X^{(\ell)}=\left[\textbf x^{(\ell)}_{1},
%\textbf x^{(\ell)}_{2},\ldots, \textbf x^{(\ell)}_{C_{\ell}} \right]\in \mathbb R^{N_{\ell}\times C_{\ell}}$.
%Similarly, we
% gather all the outputs of   $F_{\ell}$ filters  denoted by $\textbf Y^{(\ell)}\in \mathbb R^{N_{\ell}\times F_{\ell}}$, which is given by
%\begin{equation}\label{out}
%\textbf Y^{(\ell)} = \left[\textbf y_1^{(\ell)}, \textbf y_2^{(\ell)}\ldots, \textbf y_{F_{\ell}}^{(\ell)}\right].\nonumber
%\end{equation}

%The output feature map on the $\ell$-th hidden layer can be further written in a compact form as
%\begin{equation}\label{out2}
%\textbf Y^{(\ell)} = \sum_{k=0}^{K}\textbf A^k \textbf X \textbf W_k^{(\ell)} + \textbf B,
%\end{equation}
%where
%\[\textbf W_k^{(\ell) } =
%\begin{bmatrix}
%g^{(\ell)}_{1,1,k}       & g^{(\ell)}_{1,2,k}   & \dots & g^{(\ell)}_{1,F_{\ell},k} \\
%g^{(\ell)}_{2,1,k}        & g^{(\ell)}_{2,2,k}   & \dots & g^{(\ell)}_{2,F_{\ell},k} \\
%\vdots&\vdots&\dots &\vdots \\
%g^{(\ell)}_{C_{\ell},1,k}       & g^{(\ell)}_{C_{\ell},2,k}    & \dots & g^{(\ell)}_{C_{\ell},F_{\ell},k}
%\end{bmatrix},\]
%and $\textbf B=[\textbf b_1, \textbf b_2,\ldots, \textbf b_{F_{\ell}} ]$.
Following the CNN architecture, an additional  nonlinear operation, e.g,   rectified linear unit (ReLU) is used after every graph convolution operation:
$$\textbf x_{f}^{(\ell+1)} = \sigma\left(\textbf y_{f}^{(\ell)}\right), $$
where $\sigma(\cdot)$ denotes the ReLU activation function  applied to the vertex values.

\subsection{Analysis of Graph Convolutional Layers}

{In the following, we analyze the mechanisms of the graph convolutional layers. 
	We start from the graph filter with the form of a {monomial} $g_{\ell}\textbf A^{k_{\ell}}$ for the $\ell$-th layer and $C_{\ell} = F_{\ell} = 1$ with $\ell =1,\ldots L$.
	In the following, we show that when the graph convolutional layers go deeper, the output of the last graph convolutional layer is proportional to the projection of the output data of the first convolutional layer along the  eigenvector corresponding to the eigenvalue of the graph adjacency matrix with the largest amplitude.
	
	\begin{theorem}
		For any filters $\textbf A^{k_{\ell}}$, with $k_{\ell}\in \{1,2,3,\ldots\}$, 
		$$\lim_{L\to +\infty} \underbrace{\sigma\left( \sigma\cdots \sigma \left(g_2 \textbf A^{k_2} \sigma\left(g_1 \textbf A^{k_1} \textbf x \right)\right)\right)}_{\textrm{$L$ times $\sigma(\cdot)$}}
		= m
		\
		\langle \textbf y_1^{(1)}, \textbf v_1 \rangle \textbf v_1.
		$$
		with $m=\prod_{\ell=1}^{\ell = L}g_{\ell}$ and $\textbf y_1^{(1)} =\sigma\left(g_1 \textbf A^{k_1} \textbf x\right)$.
	\end{theorem}	
	
	\noindent {\textbf {Proof}} \quad For any input  data  $\textbf x\in \mathbb R^N$ on the graph with $N$ vertices, the output of the first graph convolutional layer is 
	$\textbf y_1^{(1)} = \sigma\left(g_1 \textbf A^{k_1} \textbf x\right).$
	According to the definition of ReLU function, we know that each component in $\textbf y_1^{(1)}$ is nonnegative. 
	The data feeded to the fully connected layer for classification, which is the output of  the $L$-th graph convolutional layer, is 
	$  \sigma\left( \sigma\cdots\left( \sigma\left(g_2\textbf A^{k_2}\textbf y_1^{(1)} \right)\right)\right).  $
	It can be observed that all the learned $g_i$ with $i\geq 2$ should be positive, otherwise, the output would be $\textbf 0$ resulting in an all-zero vector feeded to the fully connected layer for classification.
	Further, since all the components of $\textbf A^{k_{\ell}}$ are nonnegative and $\textbf y_1^{(1)}$ is nonnegative, by conduction,    the input of ReLU function in each layer is nonnegative. 
	Therefore, the output of the $L$-th graph convolutional layer   can be  written equivalently as 
	\begin{equation}
	\begin{split}
	&\underbrace{\sigma\left( \sigma\cdots \sigma \left(g_2 \textbf A^{k_2} \sigma\left(g_1 \textbf A^{k_1} \textbf x \right)\right)\right)}_{\textrm{$L$ times $\sigma(\cdot)$}}\\
	=& m
	\textbf A^{\sum_{\ell = 2}^{L} k_{\ell}}\textbf y_1^{(1)}\\
	{\overset{(a)}{=}} & m
	\textbf V\textbf J^{\sum_{\ell = 2}^{L} k_{\ell}}\textbf V^{-1} \textbf y_1^{(1)} \\
	{\overset{(b)}{=}}&m
	\textbf V\textbf J^{\sum_{\ell = 2}^{L} k_{\ell}}\textbf V^{-1}\left( c_1\textbf v_1 + c_2\textbf v_2\ldots c_N\textbf v_N\right)\\
	{\overset{(c)}{=}}& m
	\textbf V\textbf J^{\sum_{\ell = 2}^{L} k_{\ell}}\left(c_1\textbf e_1+c_2\textbf e_2\ldots+c_N\textbf e_N\right).
	\end{split}
	\end{equation}
	In (a), we use eigendecomposition of $\textbf A$, where $\textbf V=[\textbf v_1, \ldots, \textbf v_N]$ with $\textbf v_i$ the eigenvector of $\textbf A$, and $\textbf J$ is a diagonal matrix with diaonal elements being the eigenvalues of $\textbf A$.\footnote{When $\textbf A$ is asymmetric and  rank deficient, Jordan decomposition is adopted to replace eigendecomposition, and then $\textbf J$ is a block diagonal matrix. The remaining analysis also applies for the Jordan decomposition.}
	Equation (b) is due to the fact that 
	the set of  eigenvectors $\left\{\textbf v_i\right\}_{i=1}^N$ form an orthogonal basis, and one can express $\textbf  y_1^{(1)}\in \mathbb R^N$ by a linear combination of    those vectors, with $c_i = \langle \textbf y_1^{(1)}, \textbf v_i \rangle$.
	In (c), $\left\{\textbf e_i\right\}_{i=1}^{N}$ is the standard basis.
	
	Without loss of generality,  the graph is assumed to be strongly connected, and  we have the unique largest eigenvalue. Then,  following the definition of   $\textbf A$, we have 
	$\textbf J =\textrm{diag}\left([1, \lambda_2, \ldots, \lambda_N] \right)$  with $|\lambda_k|<1$ for all $k>2$. Then we obtain 
	\begin{equation}
	\begin{split}
	&\lim_{L\to +\infty}\textbf V\textbf J^{\sum_{\ell = 2}^{L} k_{\ell}}\left(c_1\textbf e_1+c_2\textbf e_2\ldots+c_N\textbf e_N\right)\\
	&=
	c_1\textbf v_1
	= \langle \textbf y_1^{(1)}, \textbf v_1 \rangle \textbf v_1.
	\end{split}
	\end{equation} \QEDA
	
	Note that  when $k_{\ell}=1$ for all $\ell\in \{1,2,\ldots L\}$, the graph convolutional filter reduces to $g_{\ell}\textbf A$ which is used in \cite{thomas2017semi}.
	Due to the linear approximation (projection along the  eigenvector corresponding to the largest eigenvalue amplitude), the information loss would degrade the classification accuracy.
	However, if we choose the graph filter as a set of filters from size-1 to size-$K$, it is not a projection anymore, and 
	the representation capability of graph convolutional layers is improved. }

\subsection{Filter Design for TAGCN Convolutional Layers}
{{ In this section, we would like to understand the proposed convolution
		as a feature extraction operator in traditional CNN rather than as embedding propagation. Taking this point of view helps us to profit from the design knowledge/experience from traditional CNN and apply it to grid structured data. Our definition of weight of a path and the following filter size for graph convolution in this section make it possible to design a graph CNN architecture similar to GoogLeNet \citep{googlenet}, in which a set of filters with different sizes are used in each convolutional layer. In fact, we found that a combination of size 1 and size 2 filters gives the best performance in all three data sets studied, which is a polynomial with maximum order 2. }

	In traditional CNN,  a $K\times K\times C_{\ell}$  filter scans over the input grid-structured data  for feature extraction.
	For image classification problems, the value  $K$ varies for different  CNN architectures and tasks to achieve  better performance.
	For example, in VGG-Verydeep-16 CNN model \citep{vgg}, only $3\times 3\times C_{\ell}$  filters are used; in ImageNet CNN model \citep{imagenet}, $11\times 11\times C_{\ell}$ filters are adopted; and in GoogLeNet \citep{googlenet}, rather than using the same size  filter in each convolutional layer,  different size filters, for example, $1\times 1\times C_{\ell}$, $3\times 3\times C_{\ell}$ and $5\times 5\times C_{\ell}$ filters, are concatenated in each convolution layer.
	Similarly,  we propose a general $K$-localized filter for graph CNN.
	
	For a graph-structured data, we cannot
	use a square filter window since the graph topology is no longer a grid.
	In the following, we demonstrate that the convolution operation $ \textbf G^{(\ell)}_{c,f}\textbf x^{(\ell)}_{c}$ with $\textbf G^{(\ell)}_{c,f}$ a polynomial filter
	$\textbf G^{(\ell)}_{c,f} =  \sum_{k = 0}^{K} g^{(\ell)}_{c,f,k}\textbf A^{k}$   is equivalent to using a  set of filters with filter size  from $1$ up to $K$.
	Each $k$-size filter, which is used for local feature extraction on the graph, is
	$k$-localized in the vertex domain.
	
	Define a \emph{path} of length $m$ on a graph $\mathcal G$ as a sequence $v = (v_0,v_1,...,v_m)$ of vertices $v_k \in \mathcal V $  such that
	each step of the path $(v_k,v_{k+1})$ corresponds to an (directed) edge of the graph, i.e., $(v_k,v_{k+1}) \in \mathcal {E}$.
	Here one path may
	visit the same vertex or cross the same edge multiple times.
	The following adjacency matrix $\textbf A $ is one such   example:
	\[\textbf A =
	\begin{bmatrix}
	\begin{smallmatrix}
	0      & 1   & 0 & 2&3&0&0&\cdots \\
	1     & 0   & 4 & 5&0&0&0&\cdots   \\
	0     & 1   & 0 & 0&0&0&1&\cdots   \\
	1     & 1   & 0 & 0&6&0&0&\cdots   \\
	1     &0   & 0 & 1&0&1&0&\cdots\\
	0     &0   & 0 & 0&1&0&0&\cdots\\
	0     &0   & 0 & 1&0&0&0&\cdots\\
	\vdots     &\vdots    & \vdots  & \vdots &\vdots &\vdots &\vdots &\ddots
	\end{smallmatrix}
	\end{bmatrix}.\]
	Since $\textbf A$ is asymmetric, it represents a directed graph, given  in  Fig. \ref{f2}.
	In this example, there are $6$ different length $3$-paths on the graph from vertex $2$ to vertex $1$, namely, $(2,1,4,1)$, $(2,1,2,1)$,  $(2,1,5,1)$, $(2,3,2,1)$, $(2,4,2,1)$, and $(2,4,5,1)$.
	
	We further define the
	\emph{weight of a path} to be the product of the edge weights along the path, i.e.,
	$\phi( p_{0,m}) = \prod_{k=1}^{m}\textbf A_{v_{k-1},v_k}$, where $p_{0,m}=(v_0, v_1,\ldots v_m)$.
	For example,
	the weight of the path $(2,1,4,1)$ is $1\times 1\times 2=2$.
	Then, the $(i,j)$th entry of $\textbf A^k$ in (\ref{con}), denoted by $\omega (p_{j,i}^k) $, can be interpreted as
	the sum of the weights of  all the length-$k$ paths from $j$ to $i$, which is $$\omega (p_{j,i}^k) = \sum_{j\in \left\{\tilde j|\tilde j   \textrm{ is $k$ paths to $i$}\right\} }\phi( p_{j,i}).$$
	In the above example, it can be easily verified that
	$\textbf A^3_{1,2} = 18$ by summing up
	the weights of all the above six paths  from vertex $2$ to vertex $1$ with length $3$.
	Then, the $i$th component of
	$\textbf A^k\textbf x^{(\ell)}_{c}$
	is the weighted sum of the input features of each vertex  $\textbf x^{(\ell)}_{c}$ that are length-$k$ paths  away to vertex $i$.
	Here,
	$k$ is defined as the
	\emph{filter size}.
	The output feature map is a vector with each component  given by the size-$k$ filter sliding on the graph following a fixed order of the vertex indices.
	
	The output at the $i$-th component
	can be written explicitly as
	$
	\sum_{c=1}^{C_{\ell}}\sum_{j}
	g^{(\ell)}_{c,f,k}\omega (p_{j,i}^k) \textbf x^{(\ell)}_{c}(j)$.
	This weighted sum is similar to the dot product for convolution for a grid-structured data in traditional CNNs.
	Finnaly, the output feature map is a weighted sum of convolution results from filters with different sizes, which is
	\begin{equation}\label{tagcn}
	\textbf y_f^{(\ell)}(i) = \sum_{k=1}^{K_{\ell}} \sum_{c=1}^{C_{\ell}}\sum_{j\in \{\tilde j|\tilde j   \textrm{ is $k$ paths to $i$}\} }\!\!\!\!\!\!\!\!\!\!\!
	g^{(\ell)}_{c,f,k}\omega (p_{j,i}^k) \textbf x^{(\ell)}_{c}(j) +  b_f\textbf 1_{N_{\ell}}.
	\end{equation}
	The above equation shows that
	each neuron in the graph convolutional layer is connected only to a local region (local vertices and edges) in the vertex domain of the input data volume, which is adaptive to the graph topology.
	The strength of correlation   is explicitly utilized in  $\omega (p_{j,i}^k)$.
	We refer to this method as topology adaptive graph convolutional network (TAGCN).

	In Fig.~\ref{f2}, we show TAGCN with an example of $2$-size filter sliding from vertex $1$ (figure on the left-hand-side) to vertex $2$ (figure on the right-hand-side).
	The filter is first placed at vertex $1$. Since paths  $(1,2,1)$ $(5,4,1)$ and so on (paths with red glow) are all $2$-length paths to vertex $1$, they are covered by this $2$-size filter. Since paths $(2,3)$ and  $(7,3)$ are not on any $2$-length path to vertex $1$, they are not covered by this filter.
	Further, when this 2-size filter moves to vertex $2$, paths $(1,5)$, $(4,5)$ and $(6,5)$ are no longer covered, but paths $(2,3)$ and $(7,3)$ are first time covered and contribute to the convolution with output at vertex $2$.
	
	Further, $\textbf y_f^{(\ell)}(i)$
	is the weighted sum of the input features of vertices in  $\textbf x^{(\ell)}_{c}$ that are within $k$-paths  away to vertex $i$, for $k=0, 1,\ldots K$, with weights given by the products of components of $\textbf A^k$ and $g^{(\ell)}_{c,f,k}$.
	Thus the output is the weighted sum of the feature map given by the filtered results from $1$-size up to $K$-size filters.
	It is evident  that the vertex convolution on the graph using $K$th order polynomials is $K$-paths localized.
	Moreover, different vertices on the graph share $g^{(\ell)}_{c,f,k}$.
	The above local convolution and weight sharing properties  of the convolution  (\ref{tagcn}) on a graph  are very similar to those in traditional CNN.
	
	{Though the convolution operator defined in (\ref{con}) is  defined on the vertex domain, it can also be understood as a filter  in the spectrum domain, and it is consistent with the definition of convolution in graph signal processing. We provide detailed discussion in the appendix.}
	%		\begin{figure}[t]
	%			\begin{center}
	%				\includegraphics[width=0.6\columnwidth, height = 2.2cm]
	%				{./Neura.png}
	%			\end{center}
	%			\caption{Graph convolution operation at each neuron in the graph convolution layer. }
	%			\label{f2}
	%		\end{figure}
	
	\begin{figure}[t]
		\begin{center}
			\includegraphics[width=0.8\columnwidth, height =5.5cm]
			{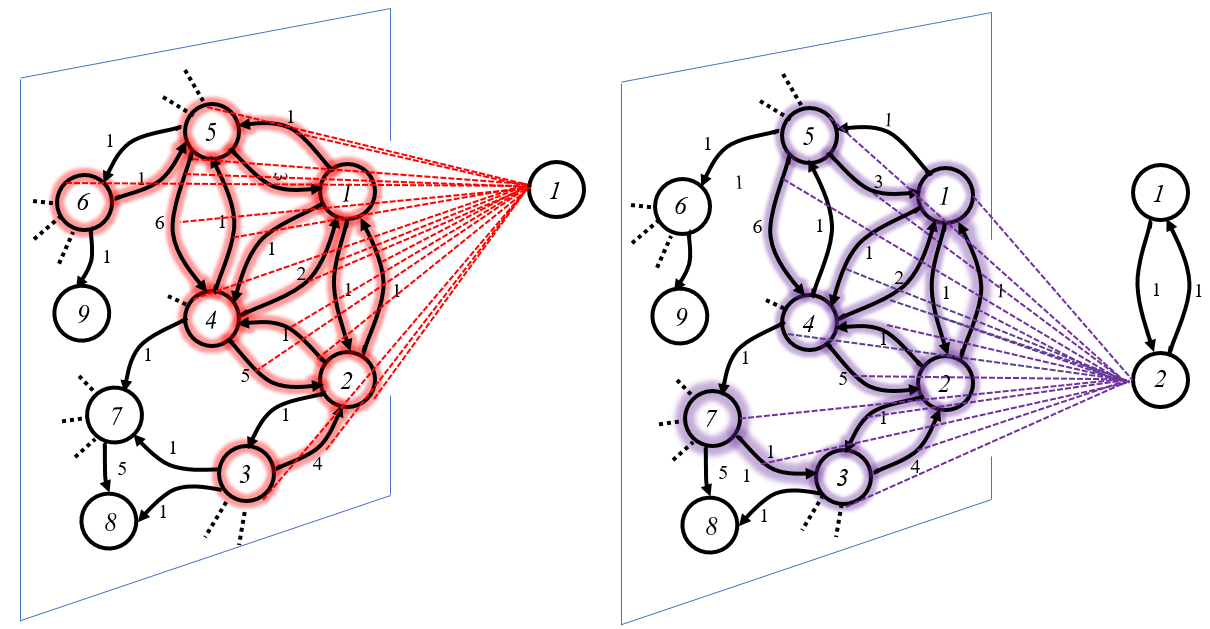}
		\end{center}
		\caption{An example of a directed graph with weights along directed edges corresponding to $\textbf A$.	
			The parts with  glow on left(right)-hand-side represent filters at different locations.
			The figure on the left-hand-side denotes the filtering/convolution starting from vertex $1$, then the filter slides  to vertex $2$  as shown on the right-hand-side with filter topology adaptive to the new local region.
			%   Note, there are multiple filters ($F_{\ell}$ in this example) along the depth, all looking at the same region covered by the
			%dashed lines that denote the convolution operation.
		}
		\label{f2}
	\end{figure}

	\section{Relation with Other Existing Formulations}
	%In this section, we show the connections and differences between the proposed TAGCN   and existing methods.
	In general, there are two types of graph convolution operators for the CNN architecture.
	One   defines the convolution  in the spectrum domain, whose output feature map is the multiplication of the inverse Fourier transform matrix with the filtered results in the spectrum domain \citep{bruna2013spectral, defferrard2016convolutional, levie2017cayleynets}.
	By doing further approximations based on this spectrum domain operator, a simplified convolution was obtained in~\citet{thomas2017semi}.
	The other defines convolution by a feature propagation model   in the vertex domain such as MoNet in ~\citet{monti2017geometric} and the diffusion CNN (DCNN) in~\cite{diffusion}.
	We investigate in detail  each alternative.

	In \citet{bruna2013spectral,defferrard2016convolutional, levie2017cayleynets}, the convolution operation was defined using the convolution theorem and  filtering operation in the spectrum domain by computing the eigendecomposition of the normalized Laplacian matrix of the graph.
	The Laplacian matrix $\textbf L$ is defined as
	$\textbf L = \textbf D - \textbf A$ with the further assumption that $\textbf A$ is symmetric to guarantee that $\textbf L$ is positive semi-definite.
	The convolution defined by the multiplication in the spectrum domain is approximated by \citet{defferrard2016convolutional} by
	\begin{equation}\label{Cheby}
	\textbf U g_{\theta} \textbf U^T\textbf x
	\approx \sum_{k=0}^{K}
	\theta_kT_k\left[\frac{2}{\lambda_{\textrm{max}}}\textbf L - \textbf I\right]
	\textbf x^{(\ell)}_{c},
	\end{equation}
	where $T_k\left[\cdot\right]$ is the $k$th order matrix Chebyshev polynomial~\citep{shuman2013emerging} where
	\begin{equation}\label{ChePoly}
	T_k(\textbf L) = 2\textbf L T_{k-1}[\textbf L] - T_{k-2}[\textbf L],
	\end{equation}
	with the initial values defined as $T_0[\textbf L]=\textbf I$ and $T_1\left[\textbf L\right] = \textbf L.$
	We refer later to this method as ChebNet for performance comparison.
	Note that the Laplacian matrix can be seen as
	a differentiator operator.
	The assumption of  symmetric  $\textbf A$ restricts the application to undirected graphs.
	Note that in \citet{defferrard2016convolutional}, Laplacian matrix polynomials with maximum order $K =25$  is needed to approximate the convolution operation on the left-hand side in (\ref{Cheby}), which imposes the computational burden.
	While TAGCN only needs an adjacency matrix polynomials with maximum order $2$ to achieve better performance as shown in the experiment part.
	
	In \citet{thomas2017semi},   a graph convolutional network (GCN) was obtained by { a first order approximation of (\ref{Cheby}).}
	In particular, let $K=1$   and  make the further assumptions that  $\lambda_{\textrm{max}} =2$ and $\theta_0 = \theta_1=\theta$.
	Then a simpler convolution operator that does not depend on the spectrum  knowledge is obtained as
	\begin{equation}\label{Cheby2}
	\textbf U g_{\theta} \textbf U^T\textbf x
	\approx \sum_{k=0}^{1}
	\theta_kT_k\!\!\left[\frac{2\textbf L}{\lambda_{\textrm{max}}} \!- \textbf I\right]\!\!
	\textbf x^{(\ell)}_{c}
	\!\approx
	\theta(\textbf I + \textbf D^{-\frac{1}{2}}
	\textbf A \textbf D^{-\frac{1}{2}} )\textbf x_c^{(\ell)}.\nonumber
	\end{equation}
	Note that $\textbf I + \textbf D^{-\frac{1}{2}}
	\textbf A \textbf D^{-\frac{1}{2}}$ is a matrix with eigenvalues in $[0,2]$.
	A renormalization trick is adopted here by letting
	$\widetilde{\textbf A} = \textbf A + \textbf I$ and $\widetilde{\textbf D}_{i,i} =  \sum_{j}\widetilde{\textbf A}_{i,j}$.
	Finally, the convolutional operator is approximated by
	\begin{equation}\label{Cheby22}
	\textbf U g_{\theta} \textbf U^T\textbf x
	\approx
	\theta \widetilde{\textbf D}^{-\frac{1}{2}}
	\widetilde{\textbf A} \widetilde{\textbf D}^{-\frac{1}{2}}
	\textbf x_c^{(\ell)}= \theta \widehat{\textbf A},
	\end{equation}
	where $ \widehat{\textbf A} =  \widetilde{\textbf D}^{-\frac{1}{2}}
	\widetilde{\textbf A} \widetilde{\textbf D}^{-\frac{1}{2}} $.
	It is interesting to observe that this method though obtained by simplifying  the spectrum method has a better performance than  the spectrum method \citep{defferrard2016convolutional}.
	The reason may be because the simplified form is equivalent to propagating vertex features on the graph, which can be seen as a special case of our TAGCN method, though there are other important  differences.
	
	As we analyzed in Section 2.2,  GCN  as in (\ref{Cheby22}) or even extending it to higher order, i.e., $\theta \widehat{\textbf A}^k$ only project the input data to the graph eigenvector of the largest eigenvalue when the convolutional layers go deeper.

	%   	\[\textbf J =
	%   \begin{bmatrix}
	%   1      &     &   &    \\
	%         & \textbf J_2(\lambda_2)   &   &      \\
	%         &     & \ddots &      \\
	%         &     &   & \textbf J_N(\lambda_N)    
	%   \end{bmatrix}.\]  
	%      	\[\textbf J_k =
	%   \begin{bmatrix}
	%   \lambda_k     & 1     &   &    \\
	%   & \lambda_k   &   \ddots&      \\
	%   &     & \ddots &     1 \\
	%   &     &   & \lambda_k   
	%   \end{bmatrix}.\]

	%   \begin{equation}
	%   \lim_{N\to +\infty} \textbf J_k^N = \textbf 0
	%   	\end{equation}
	%   	Then we have 
	%   	   \begin{equation}
	%   	\lim_{N\to +\infty} \textbf A^N\textbf y = 
	%   	\lim_{N\to +\infty} \textbf A^N\textbf y = c_1\textbf v_1
	%   	\end{equation}

	{ Our TAGCN is able to leverage information at a farther distance, but it is not a simple extension of GCN \citet{thomas2017semi}.
		First,  the graph convolution in GCN is defined as a first order Chebyshev polynomial of the graph Laplacian matrix, which is an approximation to the graph convolution defined in the spectrum domain in \citet{defferrard2016convolutional}.
		In contrast, our graph convolution is rigorously defined as multiplication by polynomials of the graph adjacency matrix; this is not an approximation, rather, it simply is filtering with graph filters as defined and as being consistent with graph signal processing.}	
	
	{Next, we show the difference between our work and the GCN method in \citet{thomas2017semi} when using 2nd order (K=2, 2 steps away from the central node) Chebyshev polynomials of Laplacian matrix.
		In the GCN paper \citet{thomas2017semi}, it has been shown that $\sum_{k=0}^{1} \theta_k T_k(\textbf L) \approx \widehat{\textbf A}$ as repeated in (\ref{Cheby22}), and $T_2[\textbf L] =2\textbf L^2$
		by the definition of Chebyshev polynomial. Then, extending GCN to the second order Chebyshev polynomials (two steps away from a central node) can be obtained from the original definition in T. Kipfâ€™s GCN \citep[eqn (5)]{thomas2017semi} as $\sum_{k=0}^{2} \theta T_k(L)= \widehat{\textbf A}   + 2\textbf L^2 -\textbf I$, which is different from our definition as in (\ref{con}). Thus, it is evident that our method is not a simple extension of GCN. We apply graph convolution as proposed from basic principles in the graph signal processing, with no approximations involved, while  both GCN in \citet{thomas2017semi} and \citet{defferrard2016convolutional} \citet{levie2017cayleynets} are based on  approximating the convolution defined  in the spectrum domain.
		In our approach, the degree of freedom is the design of the graph filter-its degree and its coefficients. Ours is a principled approach and provides a generic methodology. The performance gains we obtain are the result of capturing the underlying graph structure with no approximation in the convolution operation.   }

	\citet{simonovsky2017dynamic} proposed the edge convolution network (ECC)   to extend the convolution operator from regular grids to arbitrary graphs.
	The convolution operator is defined similarly to (\ref{Cheby2}) as
	\begin{equation}
	\textbf y_f^{(\ell)}(i)
	=
	\sum_{j\in \mathcal N(i)}  \frac{1}{\left|\mathcal {N}(i)\right|} \bm\Theta^{(\ell)}_{j,i}\textbf x^{(\ell)}_{c}(j) + b_{f}^{(\ell)},\nonumber
	\end{equation}
	with $\bm\Theta^{(\ell)}_{j,i}$ is the weight matrix that needs to be learned.

	A mixture model network (MoNet) was proposed in  \cite{monti2017geometric}, with convolution  defined as
	\begin{equation}
	\textbf y^{(\ell)}_{f} (i)
	=
	\sum_{f=1}^F\sum_{j\in \mathcal N(i)}
	g_f \kappa_f
	\textbf x^{(\ell)}_{c}(j),\nonumber
	\end{equation}
	\vspace{-0.1cm}where
	$\kappa_f$ is a Gaussian kernel with
	$\kappa_f = \exp\left\{-\frac{1}{2}(\textbf u - \bm\mu_f)^T\mathbf\Sigma_f^{-1}(\textbf  u - \bm\mu_f)\right\}$
	and
	$g_f$ is the weight  coefficient for each Gaussian kernel $\kappa_f$.
	It is further assumed that $\mathbf\Sigma_f$ is a $2\times 2$ diagonal matrix.
	
	GCN, ECC, and MoNet  all  design a propagation model on the graph; their
	differences are on the weightings used by each model.

	%then each hidden layer, there are $4F_{\ell}C_{\ell}$ coefficients need to be learned.

	\cite{diffusion} proposed a diffusion CNN (DCNN) method that considers a diffusion process over the graph.
	The transition probability of a random walk on a graph is given by $\textbf P =\textbf D^{-1}\textbf A$, which is equivalent to the normalized  adjacency matrix.
	\begin{equation}
	\textbf y_{c,f}^{(\ell)} = \textbf g^{(\ell)}_{c,f}\textbf P^f\textbf x^{(\ell)}_{c}.\nonumber
	\end{equation}
	%In the $\ell$-th hidden layer, if the input dimension is $C_{\ell}$, and $F_{\ell}$ filters are used,  the   dimension of the $\textbf Y^{(\ell)} \in N\times C_{\ell} F_{\ell}$ is  $\textbf Y^{(\ell)} = \left[\textbf y_{1,1},\ldots \textbf y_{C_{\ell}, F_{\ell}}\right]$.
	
	By comparing the above methods with TAGCN in (\ref{tagcn}), it can be concluded that
	GCN, ECC and MoNet can be seen as a special case of TAGCN because in (\ref{tagcn}) the item with $k=1$  can be seen as an  information propagation term.
	However, as shown in Section~2.2, when the convolutional layers go deeper, the output of the last convolutional layer  of GCN, ECC and MoNet are all   linear approximations of the output of the corresponding first convolutional layer which degradates the representation capability. TAGCN overcomes this  by
	designing a set of fixed size  filters that is adaptive  to the input graph topology when performing convolution on the graph.
	Further,	compared with  existing spectrum methods \citep{bruna2013spectral, defferrard2016convolutional, levie2017cayleynets}, TAGCN satisfies the convolution theorem as shown in the previous subsection and  implemented in the vertex domain, which avoids performing costly and practically numerical unstable eigendecompositions.
	
	We further compare the number of weights that need to be  learned in each hidden layer for these different methods in Table 1.
	As we show later,  $K=2$ is selected in our experiments
	using cross validation.
	However, for ChebNet in \citep{defferrard2016convolutional}, it is suggested that one needs a $25^{\textrm{th}}$ degree Chebyshev polynomial to provide a good approximation to the graph Laplacian spectrum.
	Thus we have a moderate number of weights to be learned.
	In the following, we show that our method achieves the best performance for each of those commonly used graph-structured data sets.

	\section{Experiments}
	\label{others}\begin{table}[t]
		\caption{Number of weights need to be learned for the $\ell$-th layer. }
		\label{sample-table}
		\begin{center} \fontsize{9}{9}
			\begin{tabular}{llllll}
				\multicolumn{1}{c}{\bf DCNN}  &\multicolumn{1}{c}{\bf ECC}
				&\multicolumn{1}{c}{\bf ChebNet}
				&\multicolumn{1}{c}{\bf GCN}
				&\multicolumn{1}{c}{\bf MoNet}
				&\multicolumn{1}{c}{\bf TAGCN}
				\\ \hline \\
				$F_{\ell}C_{\ell}$         &$F_{\ell}C_{\ell}$   & $25F_{\ell}C_{\ell}$  &
				$ F_{\ell}C_{\ell}$  &$4F_{\ell}C_{\ell}$  &$2F_{\ell}C_{\ell}$\\
			\end{tabular}
		\end{center}
	\end{table}
	The proposed TAGCN is general and can be fit to the general graph CNN architectures for different tasks. In the experiments, we focus on the vertex semisupervised learning problem, where we have access to only a few labeled vertices, and the task is to classify the remaining unlabeled vertices. To compare the performance of TAGCN with that of existing methods, we extensively evaluate TAGCN on three graph-structured datasets, including the Cora, Citesser and Pubmed datasets. The datasets split and experiments setting closely follow the standard criteria in \cite{yang2016revisiting}.

	%\subsection{Datasets}
	%TAGCN is evaluated on three data sets coming from citation networks: Citeseer, Cora, and Pubmed. We closely follow the data set split and experimental setup in \cite{yang2016revisiting}. Each data set consists of a certain classes of documents, yet only a few documents are labeled. The task is to classify the documents in the test set with these limited number of labels.
	In each data set, the vertices are the documents and the edges are the citation links. Each document is represented by sparse bag-of-words feature vectors, and the citation links between documents are provided. Detailed statistics of these three data sets are summarized in Table \ref{data}. It shows the number of nodes and edges that corresponding to documents and citation links, and  the number of document classes in each data set. Also, the number of features at each vertex is given. Label rate denotes the number of labeled documents that are used for training divided by the total number of documents in each data set.

	\label{others}\begin{table}[t]
		\caption{Dataset statistics, following \cite{yang2016revisiting}}
		\label{data}
		\begin{center}\fontsize{9}{9}
			\begin{tabular}{llllll}
				\multicolumn{1}{c}{\bf Dataset}  &\multicolumn{1}{c}{\bf Nodes}
				&\multicolumn{1}{c}{\bf Edges}
				&\multicolumn{1}{c}{\bf Classes}
				&\multicolumn{1}{c}{\bf Features}
				&\multicolumn{1}{c}{\bf Label Rate}
				\\ \hline \\
				Pubmed         &19,717 & 44,338 & 3&500&0.003\\
				Citeseer         &3,327 & 4,732 &6 & 3,703&0.036\\
				Cora         &2,708 & 5,429 & 7 & 1,433&0.052\\
			\end{tabular}
		\end{center}
	\end{table}
	\begin{table}
		\caption{Summary of results in terms of percentage classification accuracy with standard variance}
		\label{result}
		\begin{center}\fontsize{9}{9}
			\begin{tabular}{lllll}
				\multicolumn{1}{c}{\bf Dataset}  &\multicolumn{1}{c}{\bf Pubmed }
				&\multicolumn{1}{c}{\bf Citeseer}
				&\multicolumn{1}{c}{\bf Cora}
				%	&\multicolumn{1}{c}{\bf Nell}
				\\ \hline \\
				DeepWalk &65.3 & 43.2 &67.2\\
				Planetoid     & 77.2   &64.7 & 75.7  \\
				DCNN     & 73.0$\pm$0.5    &- & 76.8$\pm$0.6  \\
				%	ChebNet (K=2)    & 73.8     &69.6 & 81.2  \\
				ChebNet    & 74.4     &69.8 & 79.5  \\
				GCN    & 
				{{79.0}}     &{{70.3} }& 81.5 \\
				%GCN($\textbf A^2$)         &70.8(0.6) & 81.7 (0.6)& 79.1(0.4)\\
				MoNet        & 78.81$\pm$0.4 &-  & {{81.69$\pm$0.5}}  \\
				GAT  &{79.0$\pm$0.3}      &\textbf{72.5$\pm$ 0.7}   & {83.0$\pm$0.7}      \\
				TAGCN (\textrm{ours})  &\textbf{81.1$\pm$0.4}      &{71.4$\pm$ 0.5}   & \textbf{83.3$\pm$0.7}      \\
				%	TAGCN($F = 8$, $K = 2$)          &\textbf{70.1}   & \textbf{82.5}  &\textbf{80.8}     \\
			\end{tabular}
		\end{center}
	\end{table}

	\subsection{Experimental Settings}
	We construct a graph for each data set with nodes representing documents and undirected edges\footnote{We use undirected graphs here as citation relationship gives positive correlation between two documents. However, in contrast with the other approaches surveyed here, the TAGCN method is not limited to  undirected graphs if directed graphs are better suited to the applications. } linking two papers if there is a citation relationship. We obtain the adjacency matrix $\bar{\textbf A}$ with $0$ and $1$ components and further obtain the normalized  matrix $\textbf A$.
	
	In the following experiments, we design a TAGCN with two hidden layers (obtained from cross validation) for the semi-supervised node classification. In each hidden layer, the proposed TAGCN is applied for convolution, followed by a ReLU activation. $16$ hidden units (filters) are designed for each hidden layer, and dropout is applied after each hidden layer. The softmax activation function is applied to the output of the second hidden layer for the final classification. For ablation study, we evaluate the performance of TAGCN with different filter sizes from 1 to 4. To investigate the performance for different number of parameters, we also design a TAGCN with $8$ filters for each hidden layer and compare its classification accuracy with all the baselines and TAGCN with $16$ filters. We train our model using Adam \citep{Adam} with a learning rate of $0.01$ and early stopping with a window size
	of $45$. Hyperparameters of the networks (filter size, dropout rate, and number of hidden layers) are selected by cross validation.
	
	To make a fair comparison, we closely follow the same split of training, validation, and testing sets as in \cite{thomas2017semi, yang2016revisiting}, i.e.,
	$500$ labeled examples for hyperparameters  (filter size, dropout rate, and number of hidden layers) optimization  and cross-entropy error is used for  classification accuracy   evaluation.
	The performance results of the proposed TAGCN method are an  average over $100$ runs.
	
	\subsection{Quantitative Evaluations}
	
	We compare the classification accuracy with other recently proposed graph CNN methods as well as a graph embedding methods known as DeepWalk and Planetoid \cite{perozzi2014deepwalk, yang2016revisiting}. 
	The  recent published graph attention networks (GAT) \cite{Bengio18} leveraging masked self-attentional layers is also compared.
	
	The quantitative results are summarized in Table~\ref{result}. Reported numbers denote classification accuracy in percentage. Results for DeepWalk, Planetoid, GCN, and ChebNet are taken from \citet{thomas2017semi}, and results for DCNN and MoNet are taken from \citet{monti2017geometric}. All the experiments for different methods are based on the same data statistics shown in Table 2. The datasets split and experiments settings closely follow the standard criteria in \cite{yang2016revisiting,thomas2017semi}. Table~\ref{result} shows that our method outperforms all the recent state-of-the-art graph CNN methods (DCNN, ChebNet, GCN, MoNet) by obvious margins for all the three datasets. 
	
	{These experiment results 
		corroborate our analyses in Section~2.2 and Section~3 that as no approximation to the convolution is needed in TAGCN, it achieves better performance compared with spectrum approximation method such as ChebNet and GCN.
		Further, using a set of size-1 up to size-2 filters avoids the linear approximation by the simply size-1 filter (\textbf A in GCN \cite{thomas2017semi}), which further verify the efficacy of the proposed TAGCN.
		Compared with the most recent GAT method \cite{Bengio18}, our method exhibit obvious advantage for the largest dataset Pubmed.
		It should be note that GAT suffers from storage limitation in their model and not scale well for large-scale graph as explained by the authors. }
	
	For ablation study, we further compare the performance of different filter sizes from $K=1$ to $K=4$ in Table \ref{order}. It shows that the performances for filter size $K=2$ are always better than that for other filter sizes. The value $K=1$ gives the worst classification accuracy. As $K=1$ the filter is a monomial, this further validates the analysis in Section 2.2 that monomial filter results in a very rough approximation.  In Table \ref{order}, we also compare the performance of different number of filters, which reflects different number of network parameters.
	Note, we also choose filer size $K=2$ and filter number $F_{\ell} =8$ that results in the same number of network parameters as that in GCN, MoNet, ECC and DCNN according to Table~1.
	It shows that the classification accuracy using $8$ filters is comparable with that using  $16$ filters in each hidden layer for TAGCN. Moreover,  TAGCN with $8$ filters can still achieve higher accuracy than GCN, MoNet, ECC and DCNN methods.
	{ This proves that, even with a similar number of parameters or architecture, our method still exhibits superior performance than GCN. }

	{ As we analyzed and explained  in Section 2.2 and Section 3, TAGCN in our paper is not simply extending GCN \citet{thomas2017semi} to $k$-th order. Nevertheless, we implement $\textbf A^2$ and compare its performance with ours. For the data sets Pubmed, Cora, and Citeseer, the classification accuracies are $79.1 (80.8)$, $81.7(83.0)$ and $70.8 (71.2)$, where the numbers in parentheses are the results obtained with our method. Our method still achieves a noticeable performance advantage over $\textbf A^2$; in particular, we note the significant performance gain with the Pubmed database that has the largest number of nodes among these three data sets. }

	\begin{table}[t]
		\caption{ TAGCN  classification accuracy (ACC) with different parameters}
		\label{order}
		\begin{center} \fontsize{9}{9}
			\begin{tabular}{llll}
				\multicolumn{1}{c}{\bf Data Set}  &\multicolumn{1}{c}{\bf Filter Size}
				&\multicolumn{1}{c}{\bf Filter Number}
				&\multicolumn{1}{c}{\bf ACC}
				%	&\multicolumn{1}{c}{\bf Dropout rate}
				\\ \hline  \\[-1.5ex]
				Citeseer
				&1 & 16&68.9 \\
				&2 & 16&\textbf{71.4} \\
				&3 & 16&70.0 \\
				&4 & 16&69.8 \\
				&2 & \textbf{8}&\textbf{71.2} \\
				\hline  \\[-1.5ex]
				Cora
				&1 & 16&81.4 \\
				&2 & 16&\textbf{83.3} \\
				&3 & 16&82.1 \\
				&4 & 16&81.8 \\
				&2 & \textbf{8}&\textbf{83.0} \\
				\hline  \\[-1.5ex]
				Pubmed
				&1& 16&79.4 \\
				&2 & 16&\textbf{81.1} \\
				&3 & 16&80.9 \\
				&4 & 16&79.5 \\
				&2 & \textbf{8}&\textbf{80.8} \\
			\end{tabular}
		\end{center}
	\end{table}
	
	%\begin{table}[htp!]
	%	\caption{Summary of results in terms of classification accuracy in percent}
	%	\label{order}
	%	\begin{center}
	%		\begin{tabular}{lllll}
	%			\multicolumn{1}{c}{\bf Data Set}  &\multicolumn{1}{c}{\bf Filter Size}
	%			&\multicolumn{1}{c}{\bf Filter Number}
	%			&\multicolumn{1}{c}{\bf ACC}
	%			&\multicolumn{1}{c}{\bf Dropout rate}
	%			\\ \hline  \\[-1.5ex]
	%			Citeseer
	%			 &1 & 16&68.9 & 0.5\\
	%			 &2 & 16&\textbf{70.9} & 0.1\\
	%			 &3 & 16&70.0 & 0.9\\
	%%			 &2 & 8&{70.6} & 0.1\\
	%			 \hline  \\[-1.5ex]
	%			Cora
	%			&1 & 16&81.4 & 0.8\\
	%			&2 & 16&\textbf{82.5} & 0.7\\
	%			&3 & 16&82.1 & 0.5\\
	%			&4 & 16&81.8 & 0.5\\
	%			&2 & 8&{82.5} & 0.8\\
	%			\hline  \\[-1.5ex]
	%			Pubmed
	%			&1& 16&79.4 & 0.2\\
	%			&2 & 16&\textbf{81.1} & 0\\
	%			&3 & 16&80.9 & 0.1\\
	%			&4 & 16&79.5 & 0.5\\
	%		    &2 & 8&{80.8} & 0\\
	%			 \end{tabular}
	%	\end{center}
	%\end{table}

	\section{Conclusions}
	We have defined a novel graph convolutional network that rearchitects the  CNN architecture for graph-structured data.
	The proposed   method, known as TAGCN,  is adaptive to the graph topology as   the filter scans the   graph.
	Further, TAGCN inherits properties of the convolutional layer in classical CNN, i.e., local  feature extraction  and weight sharing.
	On the other hand, by the convolution theorem, TAGCN that implements in the vertex domain offers implement in the spectrum domain unifying graph CNN in both the spectrum domain and the vertex domain.
	TAGCN is  consistent with  convolution in graph  signal processing.
	These nice properties  lead to a noticeable performance advantage  in  classification accuracy on
	semi-supervised graph vertex classification problems with low computational complexity.

			\section{Appendix: Spectrum response of TAGCN}
			In  classical signal processing \citep{oppenheim1999discrete}, the convolution in the time domain is equivalent to  multiplication in the spectrum domain.
			This relationship is known as the convolution theorem.
			\citet{sandryhaila2013discrete}  showed
			that the  graph filtering defined in the vertex domain satisfies the generalized convolution theorem naturally and can also interpret spectrum filtering for both directed and undirected graphs.
			Recent work \citep{bruna2013spectral, defferrard2016convolutional} used the convolution theorem for undirected graph-structured data and designed a spectrum graph filtering.
			
			\begin{figure}[h]
				\begin{center}
					\includegraphics[width=0.4\columnwidth, height = 1.8cm]
					{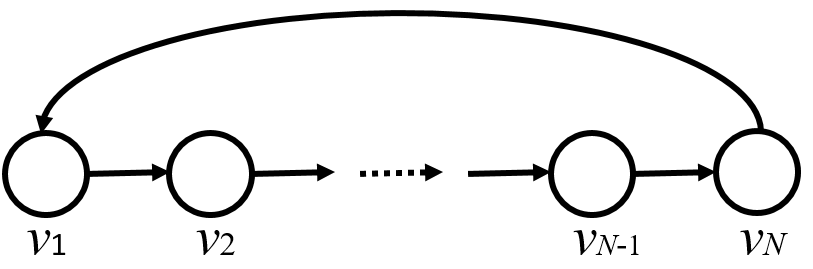}
				\end{center}
				\caption{Graph topology of a 1-D cyclic graph.}
				\label{f1}
			\end{figure}
			
			Assume that the  adjacency matrix $\textbf A$ for a  graph is diagonalizable, i.e.,
			$\textbf A = \textbf F^{-1}\textbf J \textbf F$ with $\textbf J$  a  diagonal matrix.
			The components on the diagonal of $\textbf J$ are eigenvalues of $\textbf A$, and the column vectors of $\textbf F^{-1}$ are the  right   eigenvectors of $\textbf A$; the  row vectors of $\textbf F$ are the  left   eigenvectors of $\textbf A$ \footnote{
				When $\textbf A$ is not diagonalizable,
				the columns of $\textbf F^{-1}$ and the rows  of $\textbf F$ are the generalized right and left eigenvectors of $\textbf A$, respectively.
				In this case, $\textbf F$ is no longer a unitary matrix. Also, $\textbf J$ is a block diagonal matrix; it is then in Jordan form. Interested readers may refer to \citet{sandryhaila2013discrete} or \citet{Deri} and references therein.}.
			By diagonalizing $\textbf A$ in (\ref{con}) for TAGCN, we obtain
			\begin{equation} \label{generalization}
			\textbf G^{(\ell)}_{c,f} \textbf x_c^{(\ell)}=
			\textbf F^{-1} \left( \sum_{k = 0}^{K} g^{(\ell)}_{c,f,k} \textbf J^k \right)\textbf F \textbf x_c^{(\ell)}.
			\end{equation}
			The expression on the left-hand-side of the above equation represents the filtering/convolution on the vertex domain.
			Matrix $\textbf F$   defines the graph Fourier transform \citep{sandryhaila2013discrete, SPM}, and $ \textbf F \textbf x_c^{(\ell)}$
			is the  input feature spectrum map, which is a linear mapping from the input feature on the vertex domain to the spectrum domain.
			The polynomial $ \sum_{k = 0}^{K} g^{(\ell)}_{c,f,k} \textbf J^k$ is the spectrum of the
			graph filter.
			Relation   (\ref{generalization}), which is equation (27) in \citet{sandryhaila2013discrete} generalizes   the classical convolution
			theorem to graph-structured data: convolution/filtering on the vertex domain becomes multiplication in the spectrum domain.
			When the graph is in the 1D cyclic  form, as shown in Fig. \ref{f1}, the corresponding adjacency matrix is of the form
			\[\textbf A =
			\begin{bmatrix}
			&    &  & 1 \\
			1        &     &   &   \\
			&\ddots &  &  \\
			&    &  1 &
			\end{bmatrix}.\]
			The eigendecomposition of $\textbf A$
			is
			$$\textbf A =
			\frac{1}{N} \textrm{DFT}^{-1}
			\begin{bmatrix}
			& e^{-j\frac{2\pi 0}{N}}     &  \\
			&\ddots     &      \\
			&    &   e^{-j\frac{2\pi (N-1)}{N}}
			\end{bmatrix} \textrm{DFT},
			$$
			where $\textrm{DFT}$ is the discrete Fourier transform matrix.
			The convolution operator defined in (\ref{con}) is consistent with  that in
			classical  signal processing.


\begin{thebibliography}{23}
	\providecommand{\natexlab}[1]{#1}
	\providecommand{\url}[1]{\texttt{#1}}
	\expandafter\ifx\csname urlstyle\endcsname\relax
	\providecommand{\doi}[1]{doi: #1}\else
	\providecommand{\doi}{doi: \begingroup \urlstyle{rm}\Url}\fi
	
	\bibitem[Atwood \& Towsley(2016)Atwood and Towsley]{diffusion}
	James Atwood and Don Towsley.
	\newblock Diffusion-convolutional neural networks.
	\newblock In \emph{Advances in Neural Information Processing Systems (NIPS)}.
	2016.
	
	\bibitem[Bruna et~al.(2014)Bruna, Zaremba, Szlam, and LeCun]{bruna2013spectral}
	Joan Bruna, Wojciech Zaremba, Arthur Szlam, and Yann LeCun.
	\newblock Spectral networks and locally connected networks on graphs.
	\newblock In \emph{Internatonal Conference on Learning Representations (ICLR)}.
	2014.
	
	\bibitem[Dai et~al.(2016)Dai, Dai, and Song]{dai2016discriminative}
	Hanjun Dai, Bo~Dai, and Le~Song.
	\newblock Discriminative embeddings of latent variable models for structured
	data.
	\newblock In \emph{International Conference on Machine Learning (ICML)}. 2016.
	
	\bibitem[Defferrard et~al.(2016)Defferrard, Bresson, and
	Vandergheynst]{defferrard2016convolutional}
	Micha{\"e}l Defferrard, Xavier Bresson, and Pierre Vandergheynst.
	\newblock Convolutional neural networks on graphs with fast localized spectral
	filtering.
	\newblock In \emph{Advances in Neural Information Processing Systems}, 2016.
	
	\bibitem[Deri \& Moura(2017)Deri and Moura]{Deri}
	Joya Deri and Jos{\'e} M.~F. Moura.
	\newblock Spectral projector-based graph {Fourier} transforms.
	\newblock \emph{IEEE Journal of Selected Topics in Signal Processing},
	11\penalty0 (6):\penalty0 785--795, 2017.
	
	\bibitem[Du et~al.(2016)Du, Ma, Wu, Kar, and Moura]{du2016convergence}
	Jian Du, Shaodan Ma, Yik-Chung Wu, Soummya Kar, and Jos{\'e} M.~F. Moura.
	\newblock Convergence analysis of distributed inference with vector-valued
	{G}aussian belief propagation.
	\newblock \emph{arXiv:1611.02010}, 2016.
	
	\bibitem[Du et~al.(2017)Du, Kar, and Moura]{du2016convergence}
	Jian Du, Soummya Kar, and Jos{\'e} M.~F. Moura.
	\newblock Distributed convergence verification for Gaussian belief propagation.
	\newblock \emph{accepted by
		IEEE Global Conference on Signal and Information Processing}, 2017.
	
	
	\bibitem[Glorot \& Bengio(2010)Glorot and Bengio]{glorot2010understanding}
	Xavier Glorot and Yoshua Bengio.
	\newblock Understanding the difficulty of training deep feedforward neural
	networks.
	\newblock In \emph{Proceedings of the Thirteenth International Conference on
		Artificial Intelligence and Statistics}. 2010.
	
	\bibitem[Grover \& Leskovec(2016)Grover and Leskovec]{grover2016node2vec}
	Aditya Grover and Jure Leskovec.
	\newblock node2vec: Scalable feature learning for networks.
	\newblock In \emph{Proceedings of the 22nd ACM International Conference on
		Knowledge Discovery and Data Mining (KDD)}. 2016.
	
	\bibitem[Hammack et~al.(2011)Hammack, Imrich, and
	Klav{\v{z}}ar]{hammack2011handbook}
	Richard Hammack, Wilfried Imrich, and Sandi Klav{\v{z}}ar.
	\newblock \emph{Handbook of product graphs}.
	\newblock CRC press, 2011.
	
	\bibitem[Kinga \& Ba(2015)Kinga and Ba]{Adam}
	Diederik~P. Kinga and Jimmy Ba.
	\newblock A method for stochastic optimization.
	\newblock In \emph{International Conference on Learning Representations
		(ICLR)}. 2015.
	
	\bibitem[Kipf \& Welling(2017)Kipf and Welling]{thomas2017semi}
	Thomas~N. Kipf and Max Welling.
	\newblock Semi-supervised classification with graph convolutional networks.
	\newblock In \emph{Internatonal Conference on Learning Representations (ICLR)}.
	2017.
	
	\bibitem[Krizhevsky et~al.(2012)Krizhevsky, Sutskever, and Hinton]{imagenet}
	Alex Krizhevsky, Ilya Sutskever, and Geoffrey~E. Hinton.
	\newblock Imagenet classification with deep convolutional neural networks.
	\newblock In \emph{Advances in Neural Information Processing Systems (NIPS)}.
	2012.
	
	\bibitem[LeCun et~al.(2015)LeCun, Bengio, and Hinton]{lecun2015deep}
	Yann LeCun, Yoshua Bengio, and Geoffrey~E. Hinton.
	\newblock Deep learning.
	\newblock \emph{Nature}, 521\penalty0 (7553):\penalty0 436--444, 2015.
	
	\bibitem[Levie et~al.(2017)Levie, Monti, Bresson, and
	Bronstein]{levie2017cayleynets}
	Ron Levie, Federico Monti, Xavier Bresson, and Michael~M Bronstein.
	\newblock Cayleynets: Graph convolutional neural networks with complex rational
	spectral filters.
	\newblock \emph{arXiv preprint arXiv:1705.07664}, 2017.
	
	\bibitem[Monti et~al.(2017)Monti, Boscaini, Masci, Rodol{\`a}, Svoboda, and
	Bronstein]{monti2017geometric}
	Federico Monti, Davide Boscaini, Jonathan Masci, Emanuele Rodol{\`a}, Jan
	Svoboda, and Michael~M Bronstein.
	\newblock Geometric deep learning on graphs and manifolds using mixture model
	{CNN}s.
	\newblock In \emph{Conference on Computer Vision and Pattern Recognition
		(CVPR)}. 2017.
	
	\bibitem[Oppenheim \& Schafer(2009)Oppenheim and
	Schafer]{oppenheim1999discrete}
	Alan~V. Oppenheim and Ronald~W. Schafer.
	\newblock \emph{Discrete-time signal processing (3rd Edition)}.
	\newblock Prentice Hall, 2009.
	
	\bibitem[Sandryhaila \& Moura(2013)Sandryhaila and
	Moura]{sandryhaila2013discrete}
	Aliaksei Sandryhaila and Jos{\'e} M.~F. Moura.
	\newblock Discrete signal processing on graphs.
	\newblock \emph{IEEE Transactions on Signal Processing}, 61\penalty0 (7), 2013.
	
	\bibitem[Sandryhaila \& Moura(2014)Sandryhaila and Moura]{SPM}
	Aliaksei Sandryhaila and Jos{\'e} M.~F. Moura.
	\newblock Big data analysis with signal processing on graphs: Representation
	and processing of massive data sets with irregular structure.
	\newblock \emph{IEEE Signal Processing Magazine}, 31\penalty0 (5), 2014.
	
	\bibitem[Shuman et~al.(2013)Shuman, Narang, Frossard, Ortega, and
	Vandergheynst]{shuman2013emerging}
	David~I Shuman, Sunil~K Narang, Pascal Frossard, Antonio Ortega, and Pierre
	Vandergheynst.
	\newblock The emerging field of signal processing on graphs: Extending
	high-dimensional data analysis to networks and other irregular domains.
	\newblock \emph{IEEE Signal Processing Magazine}, 30, 2013.
	
	\bibitem[Simonovsky \& Komodakis(2017)Simonovsky and
	Komodakis]{simonovsky2017dynamic}
	Martin Simonovsky and Nikos Komodakis.
	\newblock Dynamic edge-conditioned filters in convolutional neural networks on
	graphs.
	\newblock In \emph{Conference on Computer Vision and Pattern Recognition
		(CVPR)}. 2017.

	\bibitem[Jian \& Wu(2013)Jian and Wu]{Jian1}
	Jian Du and Y.-C. Wu.
	\newblock Distributed clock skew and offset estimation in wireless sensor networks: Asynchronous algorithm and convergence analysis.
	\newblock \emph{IEEE Transactions on Wireless Communications}, 12\penalty0 (11), 2013.
	
	\bibitem[Simonyan \& Zisserman(2015)Simonyan and Zisserman]{vgg}
	Karen Simonyan and Andrew Zisserman.
	\newblock Very deep convolutional networks for large-scale image recognition.
	\newblock In \emph{Internatonal Conference on Learning Representations (ICLR)}.
	2015.
	
		\bibitem[Jian \& Wu(2013)Jian and Wu]{Jian1}
	Jian Du and Y.-C. Wu.
	\newblock Network-wide distributed carrier frequency offsets estimation and compensation via belief propagation.
	\newblock \emph{IEEE Transactions on Wireless Communications}, 61\penalty0 (23), 2013.
	
	

	\bibitem[Szegedy et~al.(2015)Szegedy, Liu, Jia, Sermanet, Reed, Anguelov,
	Erhan, Vanhoucke, and Rabinovich]{googlenet}
	Christian Szegedy, Wei Liu, Yangqing Jia, Pierre Sermanet, Scott Reed, Dragomir
	Anguelov, Dumitru Erhan, Vincent Vanhoucke, and Andrew Rabinovich.
	\newblock Going deeper with convolutions.
	\newblock In \emph{Conference on Computer Vision and Pattern Recognition
		(CVPR)}. 2015.
	
	\bibitem[Yang et~al.(2016)Yang, Cohen, and Salakhutdinov]{yang2016revisiting}
	Zhilin Yang, William Cohen, and Ruslan Salakhutdinov.
	\newblock Revisiting semi-supervised learning with graph embeddings.
	\newblock In \emph{Internatonal Conference on Learning Representations (ICLR)}.
	2016.
	
\end{thebibliography}
\end{document}